\documentclass[10pt,twocolumn,letterpaper]{article}

\usepackage{cvpr}
\usepackage{times}
\usepackage{epsfig}
\usepackage{graphicx}
\usepackage{amsmath}
\usepackage{amssymb}

\usepackage[pagebackref=true,breaklinks=true,letterpaper=true,colorlinks,bookmarks=false]{hyperref}

\cvprfinalcopy 

\def\cvprPaperID{4171} 

\begin{document}

\title{Exploring the Bounds of the Utility of Context for Object Detection}

\author{Ehud Barnea and Ohad Ben-Shahar\\
	Dept. of Computer Science, Ben-Gurion University\\
	Beer-Sheva, Israel\\
	{\tt\small \{barneaeh, ben-shahar\}@cs.bgu.ac.il}
}

\maketitle

\begin{abstract}
   The recurring context in which objects appear holds valuable information that can be employed to predict their existence. This intuitive observation indeed led many researchers to endow appearance-based detectors with explicit reasoning about context. The underlying thesis suggests that stronger contextual relations would facilitate greater improvements in detection capacity. In practice, however, the observed improvement in many cases is modest at best, and often only marginal. In this work we seek to improve our understanding of this phenomenon, in part by pursuing an opposite approach. Instead of attempting to improve detection scores by employing context, we treat the utility of context as an optimization problem: to what extent can detection scores be improved by considering context or any other kind of additional information? With this approach we explore the bounds on improvement by using contextual relations between objects and provide a tool for identifying the most helpful ones. We show that simple co-occurrence relations can often provide large gains, while in other cases a significant improvement is simply impossible or impractical with either co-occurrence or more precise spatial relations. To better understand these results we then analyze the ability of context to handle different types of false detections, revealing that tested contextual information cannot ameliorate localization errors, severely limiting its gains. These and additional insights further our understanding on where and why utilization of context for object detection succeeds and fails.

\end{abstract}

\section{Introduction}
\label{sec:intro}

\begin{figure*}
	\begin{center}
		\begin{tabular}{cccc}
			\hspace*{-0.3cm}\raisebox{-0.9mm}[0pt][0pt]{\includegraphics[width=0.23\linewidth]{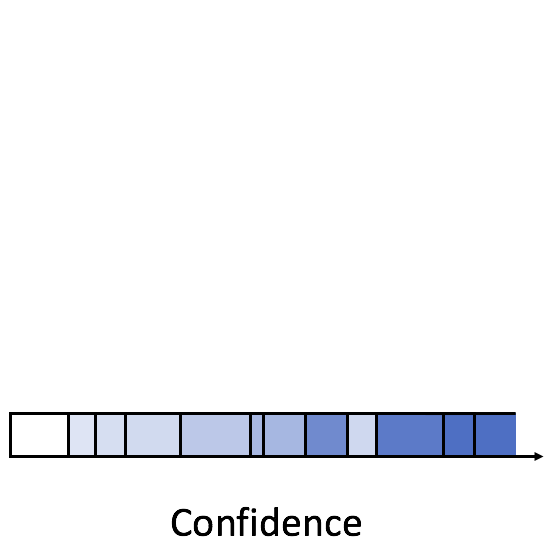}} &
			\hspace*{-0.0cm}\includegraphics[width=0.23\linewidth]{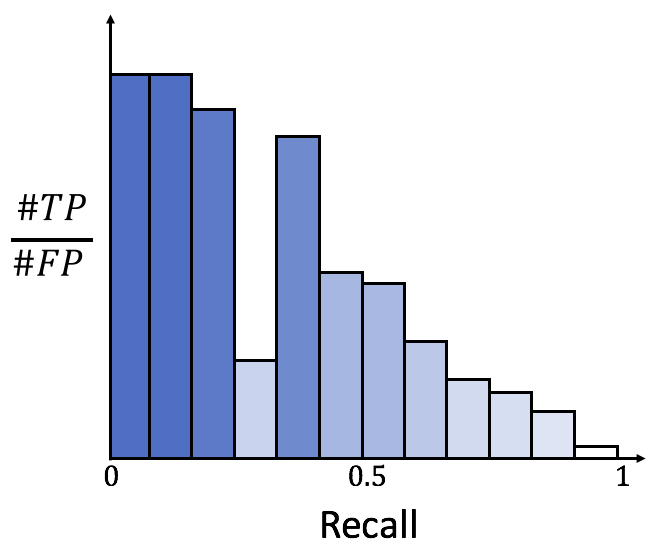} &
			\hspace*{-0.4cm}\includegraphics[width=0.23\linewidth]{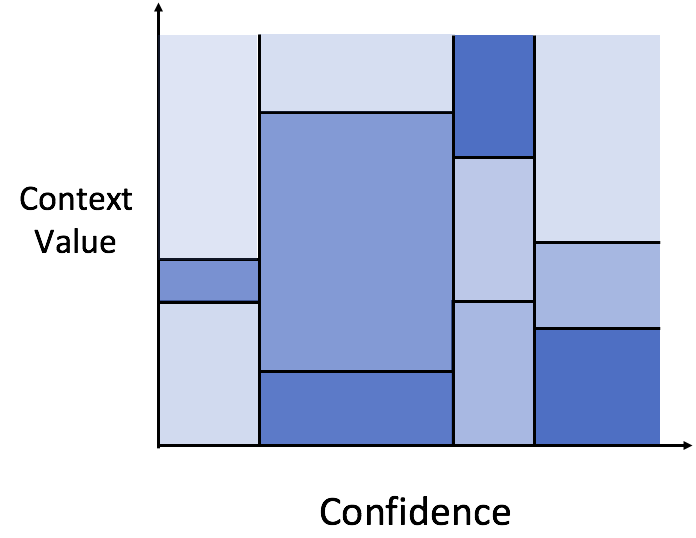} &
			\hspace*{-0.0cm}\includegraphics[width=0.23\linewidth]{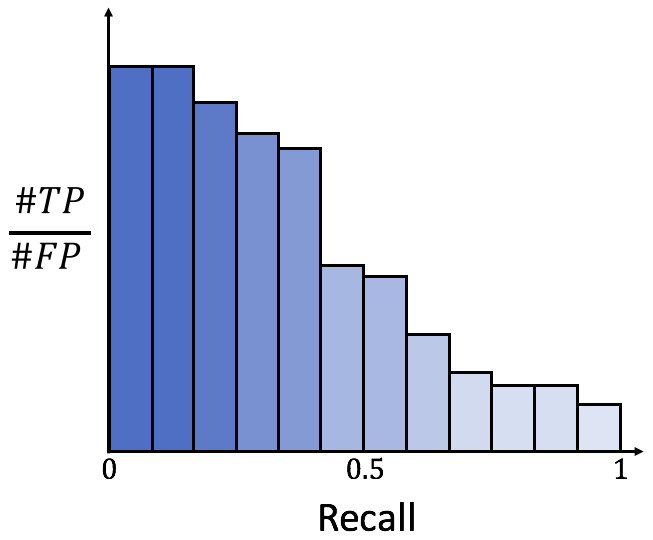} \\
			(a)&(b)&(c)&(d) \\
		\end{tabular}
	\end{center}
	\caption{Ranking detections for the evaluation of detection results (darker colors represent higher ratios $\frac{\#TP}{\#FP}$ of true to false detections in each bin). To evaluate detections by average precision (AP), the confidence space is first discretized into bins with equal number of true detections (a), which are then sorted by decreasing confidence (b) for the calculation of precision at different recall levels. When also employing context, the space of confidence and context is similarly discretized into bins with equal number of true detections (c), and a criterion for their ranking is required. We show that the non-parametric function over this space that ranks bins according to their ratio of true to false detections, as shown in (d), provides the maximal AP.}
	\label{fig:summary}
\end{figure*}

Historically, object detection has  been performed by ``sliding'' a window over an image and deciding whether it contains an object by reasoning about its appearance inside the window \cite{Felzenszwalb_McAllester_2008_CVPR,Dalal_Triggs_2005_CVPR}. Naturally, this type of calculation only takes the object's \emph{local} context into account, as more distant context falls outside of the window. Contemporary detectors based on convolutional neural networks are able to expand their reasoning beyond local context since the receptive field of neurons grows with depth, eventually covering the entire image. However, the extent to which such a network is able to incorporate context is still not entirely understood \cite{Luo_etal_2016_NIPS}.
To improve both types of detectors, many works have sought to explicitly combine their results with contextual reasoning to strengthen detections that appear in favorable context and to weaken detections that do not (among others, see \cite{divvala_etal_2009_CVPR,Bell_etal_2016_CVPR,Li_etal_2017_TransOnMulti,Heitz_Koller_2008_ECCV,Oramas_De-Raedt_Tuytelaars_2013_ICCV,Wolf_Bileschi_2006_IJCV,Torralba_Sinha_2001_ICCV,Cinbis_Sclaroff_2012_ECCV,Mottaghi_etal_2014_CVPR,Desai_etal_2011_IJCV,Torralba_etal_2004_NIPS,Perko_Leonardis_2010_CVIU,Hoiem_etal_2008_IJCV,Arbel_etal_2017_arxiv}). 

These approaches have shown significant gains in some cases \cite{Mottaghi_etal_2014_CVPR}, but in many others the explicit application of contextual information has shown negligible improvement (or even diminished results)  \cite{Yu_etal_2016_BMVC,yao_feifei_2010_CVPR}. This problem was first discussed by Wolf and Bileschi \cite{Wolf_Bileschi_2006_IJCV}, showing that context learned from low-level features can be used \emph{instead} of an appearance-based detector, but provides little aid when combined with an appearance based detector. Attempting to explain this behavior, they observed that only few samples were highly confident while being out of context. However, we are still left with no insights as to when context can be expected to improve and whether the lack of improvement observed in many cases should be attributed to (1) the limited capacity of extracting contextual information, (2) the contextual relations employed, (3) the difficulty of combining it with appearance-based confidence, or perhaps, (4) that contextual information is simply redundant once appearance information has been accounted for.

To investigate this matter we suggest a novel approach to compute an upper bound on the improvement that can be obtained by combining detections' base confidence with different contextual relations or any other kind of additional information. In this formulation a detection's context can be represented as a real number that indicates, for example, the extent to which the context supports the existence of an object. Alternatively, it may be binary, separating detections into those for which a contextual relation holds and those for which it does not (\eg, the relation ``is next to a person'' separates detections into two groups). Applying this method we are able to identify the contextual relations that have the largest improvement bounds, and so a greater potential for improving detection results. Similarly, we identify object categories for which no relevant contextual relation provides a large improvement bound. For these categories the context is meaningless even if it is extracted accurately and perfectly combined with the confidence provided by the base detector. 

To better understand why employing some relations fails to improve, we show which types of false detections can be corrected by use of context. Specifically, we analyze the ability of contextual relations to distinguish between true detections and different types of false detections (\ie localization errors, confusion between categories, and confusion with the background \cite{hoiem_etal_2012_ECCV}). This analysis reveals that while context in the form of object relations can identify confusions with the background and between categories, it cannot distinguish between true detections and localization errors in most cases, often rendering it useless. This inability to handle localization errors means that strengthening detections in favorable context also strengthens localization errors, an observation that may explain why context sometimes hurts detection results. 

Taken together, the aspects studied in our paper tell a story of detection with context. In a few cases the context can provide very significant improvements, in most other cases it can provide meaningful improvements if used correctly, while in some cases the context is simply not informative (cannot differentiate true and false detections of any type). In all cases, when more accurate localization is required the benefits of employing context are reduced. The goal of this analysis paper is to further our understanding regarding the role of context and provide a tool for identifying the most promising contextual relations, the maximal improvement they can be expected to provide, and the object categories that will be affected the most/least (if at all).

\section{Background and Survey of Detection with Context}
\label{sec:survey}

Various contextual relations have been employed for object detection, ranging from local context just outside a detection's window \cite{Felzenszwalb_etal_2010_PAMI}, to more global information such as object co-occurrence \cite{Galleguillos_etal_2008_CVPR} or spatial relations between objects \cite{Oramas_De-Raedt_Tuytelaars_2013_ICCV} or surfaces \cite{Hoiem_etal_2008_IJCV}. Different models have been suggested for combining such context with appearance-based scores, employing probabilistic models \cite{Galleguillos_etal_2008_CVPR}, discriminative classifiers \cite{Cinbis_Sclaroff_2012_ECCV}, and lately also recurrent neural networks (RNNs) \cite{Bell_etal_2016_CVPR} and neural attention models \cite{Li_etal_2017_TransOnMulti}. Since this work focuses on model results rather than methodology, we refer the readers to Mottaghi \etal \cite{Mottaghi_etal_2014_CVPR}.

Despite the repeated attempts, dissatisfaction with presented results was expressed in various cases.  Yao and Fei-Fei \cite{yao_feifei_2010_CVPR} suggested that ``most of the context information has contributed relatively little to boost performances in recognition tasks'' and suggested the lack of strong context as a reason (thus choosing to work on a different problem that relies critically on context). Choi \etal \cite{choi_etal_2010_CVPR} claimed that the PASCAL dataset \cite{Everingham_etal_2010_IJCV} is not suitable to test context-based object recognition algorithms as most of its images contain only a single object class. However, when the more elaborate SUN 09 dataset \cite{choi_etal_2010_CVPR} was employed, the improvement after including context was similar in both cases (a 1.05 increase in AP over PASCAL and 1.31 over SUN09). 

In a study of context selection, Yu \etal \cite{Yu_etal_2016_BMVC} noted that ``the performance gain from context itself has not been significant'' in previous works. Yu showed that contextual relations can predict object locations and that some relations are more predictive than others. However, these observations do not attest to the utility of context when combined with a reasonably good detector, which was shown to be negligible due to a shortage of confident false detections that were out of context \cite{Wolf_Bileschi_2006_IJCV}.

To better understand the state of context models we studied the results presented in 12 papers \cite{chu_and_dend_2016_context,Bell_etal_2016_CVPR,Li_etal_2017_TransOnMulti,choi_etal_2010_CVPR,Yu_etal_2016_BMVC,divvala_etal_2009_CVPR,chen_etal_2013_CVPR,Heitz_Koller_2008_ECCV,Mottaghi_etal_2014_CVPR,Cinbis_Sclaroff_2012_ECCV,Desai_etal_2011_IJCV,yao_etal_2012_CVPR} over different detection datasets (PASCAL VOC \cite{Everingham_etal_2010_IJCV}, SUN 09 \cite{choi_etal_2010_CVPR}, SUN RGB-D \cite{song_etal_2015_CVPR}, and MSRC-21 \cite{shotton_etal_2008_CVPR}). 
We collected results from papers in which the base detector's AP ($AP_{base}$) and the AP after including context outside a detection window ($AP_{context}$) were reported separately for each category and presented without additional algorithmic improvements that were not related to context. 
We then calculated all improvements due to the consideration of context by taking the difference $AP_{context}-AP_{base}$. These improvements (and occasional setbacks) are summarized below while the results of each individual method are provided in the supplementary material. 

\begin{figure}
	\begin{center}
		\includegraphics[width=1\linewidth]{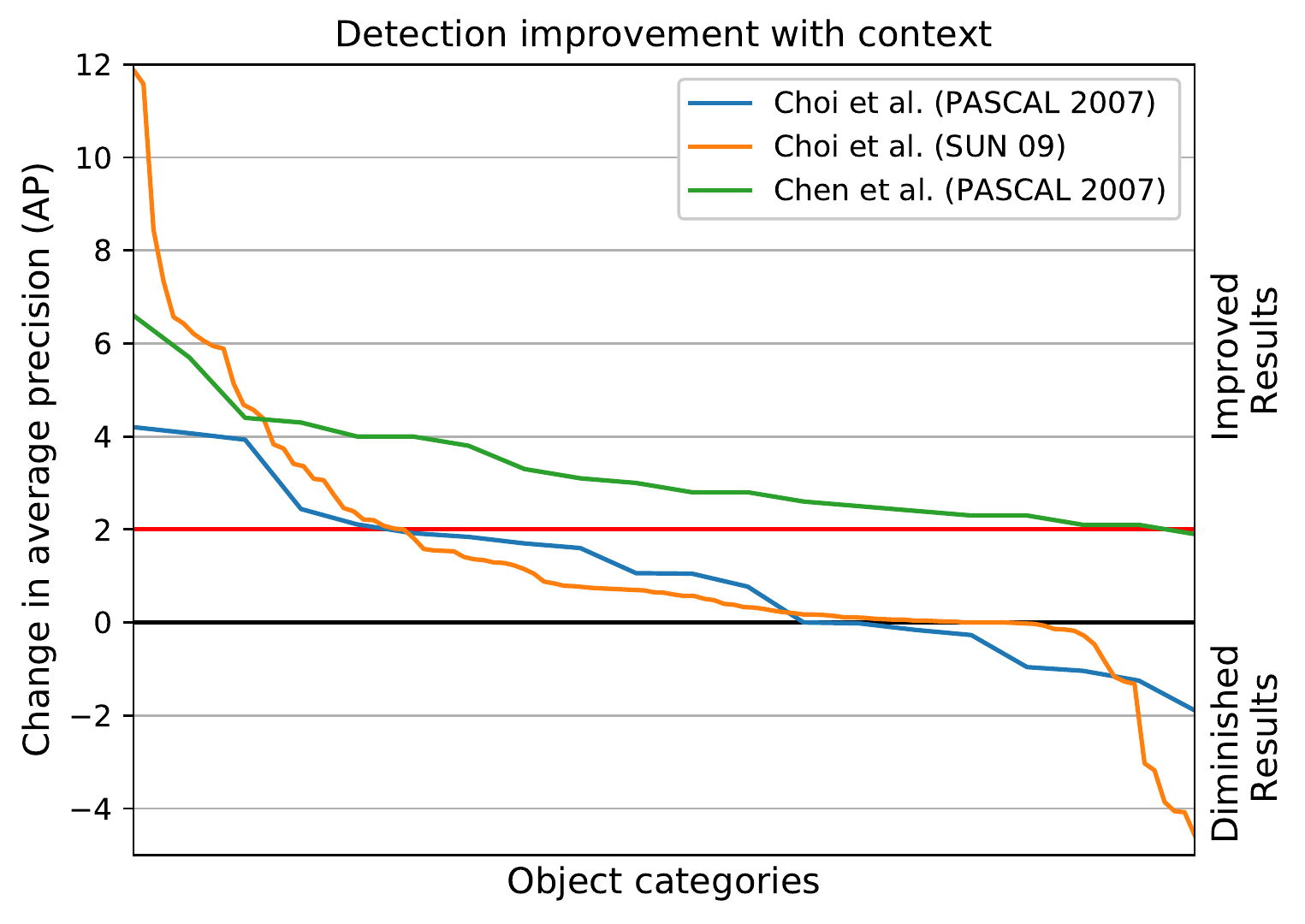} 
	\end{center}
	\caption{Improvement (or change) in detection results per category following the incorporation of context. For each method the change in average precision per object category is reported (where categories are sorted within method and thus may be ordered differently along the X axis in each method or dataset). Most methods follow the pattern of Choi \etal \cite{choi_etal_2010_CVPR}, in which the change is sometimes positive and substantial (above the red line), or positive but less substantial (below the red line), and it can also be negative.
	}
	\label{fig:litres}
\end{figure}

Examining the different methods, most papers display a pattern similar to the method of Choi \etal \cite{choi_etal_2010_CVPR} over PASCAL (Fig. \ref{fig:litres} blue). In particular, some object categories exhibit a marked improvement, some show only marginal improvement  (which is defined in this work as an increase of less than 2 AP units \footnote{Of course, there is no clear amount of improvement that is considered substantial. We believe that a 2 AP threshold would seem reasonable to many in the community and yet stress that it is indeed subjective and that none of the conclusions presented in this paper hinge on a hard distinction between results above and below this threshold.}), and others suffer from \emph{reduced} detection results. 
Fig. \ref{fig:litres} shows that applying the same method to the SUN 09 dataset (that contains more objects and so intuitively has more contextual information \cite{choi_etal_2010_CVPR}) results in a similar pattern of results (orange curve), albeit with greater variance (i.e., larger improvements but also greater diminished results). 
That being said, selected works have shown a different pattern of results with a marked improvement in most object categories \cite{Cinbis_Sclaroff_2012_ECCV,Mottaghi_etal_2014_CVPR,Heitz_Koller_2008_ECCV}, or even in all or them (green curve) \cite{chen_etal_2013_CVPR}.

\begin{figure}
	\begin{center}
		\includegraphics[width=1\linewidth]{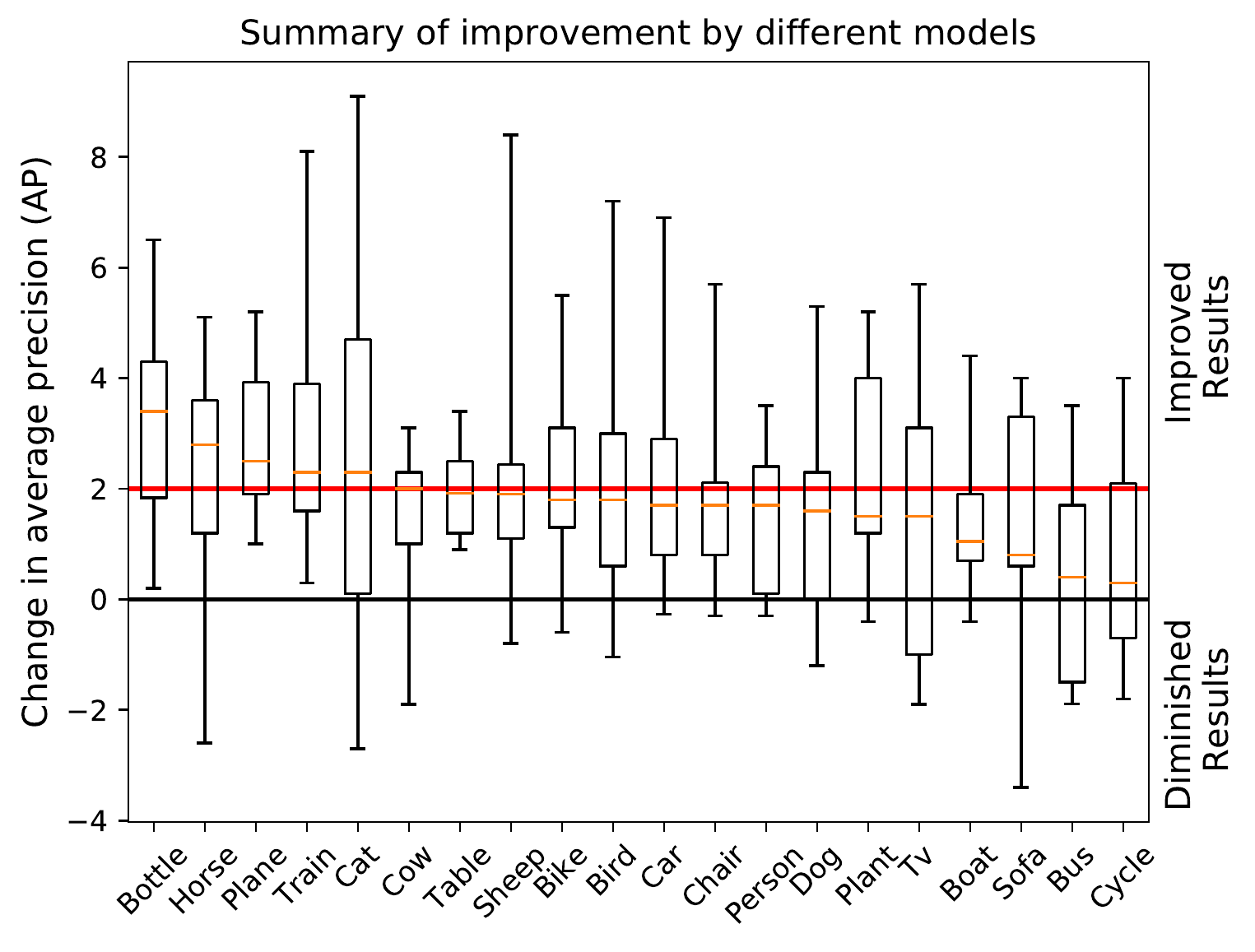}
	\end{center}
	\caption{Box plot of the improvement (or change) in detection results measured by average precision following the incorporation of context by different methods over the PASCAL dataset. Orange bars inside the boxes represent the median of methods' context-related change per category. Box top and bottom edges, as well as the ends of extending lines, all represent different quartiles. As can be seen (see also supplementary materials), only few object categories are significantly improved by most methods (i.e., more than 2 AP units, or above the red line).
	}
	\label{fig:litres_per_cat}
\end{figure}

To show the utility of context per object category, we summarize the improvement of different algorithms over PASCAL 2007-2010 in Fig. \ref{fig:litres_per_cat}. As shown, while substantial improvements were obtained for most methods on a few object categories (bottle, horse, plane, train, and cat), for most categories the majority of methods obtained an improvement of less than 2 AP units. However, for each category there exists a method that achieves a significant increase in detection results. Thus, while a substantial improvement is possible regardless of category, obtaining such general behavior by applying a single method is rather challenging.

\section{Bounding the Utility of Context}
\label{sec:bounding}

In this section we present a method for finding the best non-parametric function that combines the confidence scores of appearance based detectors and additional information (in this case: context) so as to maximize $AP$. We assume that we are given a set of detections $\{x_1,...,x_n\}$, where each detection $x_i=(\mathit{conf_i},label_i,loc_i,ctx_i)$ is defined by its location $loc_i$, object class label $label_i$, base-confidence $\mathit{conf_i}$ assigned by a base detector, and a value $ctx_i$ that represents the detection's context. $ctx_i$ can be a boolean value, indicating whether some binary contextual relation holds for $x_i$, \eg, it can indicate whether there is a person to the left of detection $x_i$. Alternatively, it can be a real value indicating the extent that the context of $x_i$ supports its label assignment $label_i$. Formally, $ctx_i$ can represent any kind of additional information about $x_i$ that we may wish to employ in order to improve detection results. 

We formulate the problem as context-based re-scoring of detections similar to most previous models \cite{divvala_etal_2009_CVPR,Cinbis_Sclaroff_2012_ECCV,Arbel_etal_2017_arxiv,Yu_etal_2016_BMVC} and define a context model as a function $g$ that calculates a new score $\mathit{conf'_i}$ for each detection $x_i$:
\begin{align}
	\label{eq:func}
	\mathit{conf'_i}=g(\mathit{conf_i},ctx_i)    \,\,   .
\end{align}
We begin by describing the calculation of AP based on confidence bins (depicted in Fig. \ref{fig:summary}a,b) and the non-parametric function $g$ that maximizes it given detections and their context, exemplified in Fig. \ref{fig:summary}c,d. At first, $ctx_i$ is assumed to be real valued and for which not many detections share the same output, as can generally be expected from standard parametric functions on image data. Later, we handle the general case in which many detections can be assigned with the exact same values $ctx_i$. This allow us to experiment with binary contextual relations based on ground-truth information, but as we show, the same definition of $g$ does not formally maximize AP but can be considered as a heuristic. Following this, we experiment with different co-occurrence and spatial contextual relations and state of the art detectors, showing the largest possible improvement in AP by employing each relation, and then provide evidence that the suggested heuristic accurately approximates the maximal AP in the general case.

\subsection{Bin-based representation of AP}
\label{sec:ap}

For a given object category, the performance of a detector is commonly evaluated using the average precision (AP) metric calculated for the detector's ranked output detections. The AP metric summarizes the shape of the precision/recall curve, depicting the detector's precision at each level of recall. Specifically, precision is defined as the fraction of all detections above a given rank which are from the positive class, and recall is defined as the fraction of positive detections that are ranked above that rank. While the AP represents an average of precision values, slightly different methods have been used for its calculation \cite{everingham_etal_2015_IJCV}. In PASCAL VOC 2007, it was defined as the mean precision at a set of eleven equally spaced recall levels [0, 0.1,..., 1]:
\begin{align}
	AP = \frac{1}{11}\sum_{i=1}^{11} p_i    \,\,   ,
\end{align}
with interpolation of precision values $p_i$. We note that while AP is in the range $[0,1]$, in this text we report AP percentage values between $[0,100]$ for clarity.

The calculation of AP is based on a ranking of detections according to their confidence, where detections are ranked higher when the detector is more confident. In practice, detections are ranked by sorting them according to \emph{decreasing} confidence. Iterating over the sorted detections from start to end, the recall at each rank gradually increases by $\frac{1}{POS}$ with every true detection encountered, where $POS$ is the number of positive examples or objects. Thus, an example recall value of $\frac{1}{10}$ corresponds to some confidence such that the number of more confident detections is a tenth of $POS$, or $\frac{POS}{10}$. Similarly, a recall of $\frac{1}{m}$ corresponds to $\frac{POS}{m}$ true detections, a recall of $\frac{2}{m}$ corresponds to $2\frac{POS}{m}$ true detections, and so on. Therefore, considering $m$ equally spaced recall levels [$\frac{1}{m}, \frac{2}{m},..., \frac{m}{m}$], each level corresponds to $\frac{POS}{m}$ true detections in addition to the previous level.

We follow a procedure similar to PASCAL 2007 without interpolation, and define the AP as the mean precision at a set of $m$ recall levels [$\frac{1}{m}, \frac{2}{m},..., \frac{m}{m}$]:
\begin{align}
	\label{eq:ap1}
	AP = \frac{1}{m}\sum_{i=1}^{m} p_i    \,\,   . 
\end{align}
These $m$ increasing recall levels correspond to $m$ \emph{decreasing} confidence values $c_1,...,c_m$ that discretize the practical confidence range $[c_m,\infty)$, into $m$ confidence bins. Each confidence bin contains an equal number of true detections $t=\frac{POS}{m}$. We note that the confidence values are not necessarily equally spaced, and $p_i=0$ is set for recall levels that cannot be obtained due to missed objects. This discretization is depicted in Fig. \ref{fig:summary}a and the corresponding recalls in Fig. \ref{fig:summary}b. Let us denote by $t_i,f_i$ the number of true and false detections in bin $i$. The AP can now be similarly represented according to the confidence bins:
\begin{align}
	\label{eq:ap2}
	AP = \frac{1}{m} \sum_{i=1}^{m} \frac{\sum_{j=1}^{i} t_j}{\sum_{j=1}^{i} (t_j  +  f_j)  }    \,\,   ,
\end{align}
and since bins have equal values of $t_i=t$:
\begin{align}
	\label{eq:ap3}
	AP = \frac{1}{m} \sum_{i=1}^{m} \frac{it}{it  +  \sum_{j=1}^{i} f_j}    \,\,   .
\end{align}

\subsection{Real-valued representation of context}
\label{sec:cont}

A non-parametric function $g: \mathbb{R} ^2 \rightarrow \mathbb{R}$ defined over detections' confidence and context (as in Eq. \ref{eq:func}) discretizes its input domain into $2D$ bins and assigns new confidence values to detections according to the input bin in which they fall. The definition of a non-parametric function requires to define this discretization, \ie, the bounding values of its bins. Since the calculation of AP discretizes the confidence space into $m$ bins, we similarly discretize the range of context values into $m$ bins of equal number of true detections $\frac{POS}{m}$. This way, the bins of $g$ coincide with those of $AP$. To do so, we first discretize the confidence axis into $m_1$ bins with $\frac{POS}{m_1}$ positives, and then similarly discretize the context axis \emph{for each confidence bin} into $m_2$ bins, each with $\frac{POS}{m_1m_2}$ positives. The total number of bins in this case is $m=m_1m_2$. The result of this process is exemplified in Fig. \ref{fig:summary}c.

Following this discretization, the definition of $g$ requires to assign a new confidence value to each bin (or rather, to detections inside it). The calculation of AP for the results of $g$ begins with the ranking of detections according to this new confidence. Because of this, what is important are not the exact values assigned by $g$, but rather the ranking they induce, or more specifically, the ranking of $g$'s $m$ bins. Thus, the question is which bin ranking provides the maximal AP and how to find it.

The bins in Eq. \ref{eq:ap3} are ordered by increasing recall. As can be seen, bins ordered by increasing number of false detections $f_i$ maximize AP since the divisors of some summands will become larger by switching $f_i$ terms with larger $f_j$ terms, thus decreasing AP. Therefore, we define the new confidence assigned to detections in bin $i$ to be $\frac{t_i}{f_i}$, as shown in Fig. \ref{fig:summary}c,d. Since $t_i$ is equal for each $i$, the confidence decreases when the number of false detections increases, providing the ranking that maximizes AP.

It is important to note that the presented method maximizes AP for a given discretization. More complicated binning schemes for $g$ may also have an equal number of true detections while providing a larger $AP$, but in this work we chose to consider a more standard discretization scheme. Finally, we note that the parameters $m_1,m_2$ that form $m$ are predetermined according to the wanted number of recall bins in the calculation of AP. Choosing $m=11$ corresponds to a calculation similar to PASCAL 2007 that employs 11 recall bins, and the maximal $m$ corresponds to newer PASCAL versions that use a recall bin for each true detection ($m=POS$).

\subsection{General representation of context}
\label{sec:general_case}

In cases when the context value of several true detections is the same they may not be possible to divide with a threshold. A discretization into bins with equal number of true detections cannot be ensured in such cases, as well as an accurate calculation of AP. In practice, when only few true detections are inseparable the results may not be highly affected, but this may become problematic when many such true detections exist.  While such a result is probably unlikely with the real-valued context functions often used in previous works and in Sec. \ref{sec:cont}, it indeed occurs when using binary contextual relations as we use for experiments in Section \ref{sec:bounding_exp}.

In this section we handle the general case of non-parametric functions $g$ under arbitrary discretization, in which the number of true detections in each bin may vary. The formulation of AP in Eq. \ref{eq:ap2} and Eq. \ref{eq:ap3} can be seen as an approximation of the area under the precision/recall curve as a Riemann sum with rectangles of heights $p_i$ and equal width $\frac{1}{m}$. As a generalization, we consider an approximation based on a Riemann sum with rectangles of heights $p_i$ and different widths $\Delta r_i$. For recall levels $[r_1,...,r_m]$ that are not necessarily equally spaced, we define $\Delta r_i=(r_i-r_{i-1})$ with $\Delta r_1=r_1$ and represent AP as:
\begin{align}
	\label{eq:ap_general}
	AP = \sum_{i=1}^{m} p_i \Delta r_i   \,\,   .
\end{align}


The new confidence of $\frac{t_i}{f_i}$ suggested in Sec. \ref{sec:cont} does not generally maximize this AP. For example, we consider a function with three bins such that $t_1=277$, $t_2=371$, $t_3=69$, and $f_1=16$, $f_2=955$, $f_3=178$, when the number of all positives is $POS=t_1+t_2+t_3$. The assigned confidence values in this case are $17.3,0.3884,0.3876$, respectively, and the ranking this induces provides an AP of 60.9. However, switching the confidence values of the second and third bins provides a ranking with a larger AP of 62.6. Therefore, in the general case we consider this new confidence as a heuristic for the maximization of AP. While a maximal AP is not always achieved, we show in Section \ref{sec:bounding_exp} that it likely provides an accurate approximation of AP. 

\subsection{Exploring the relations between objects}
\label{sec:bounding_exp}

Using this framework, we can now examine the upper bound of AP for different co-occurrence and spatial contextual relations where a single relation is considered in each experiment. We then explore the relations that provide the highest and lowest AP upper bounds per object category and present various insights.

\begin{figure}
	\begin{center}
		\includegraphics[width=1\linewidth]{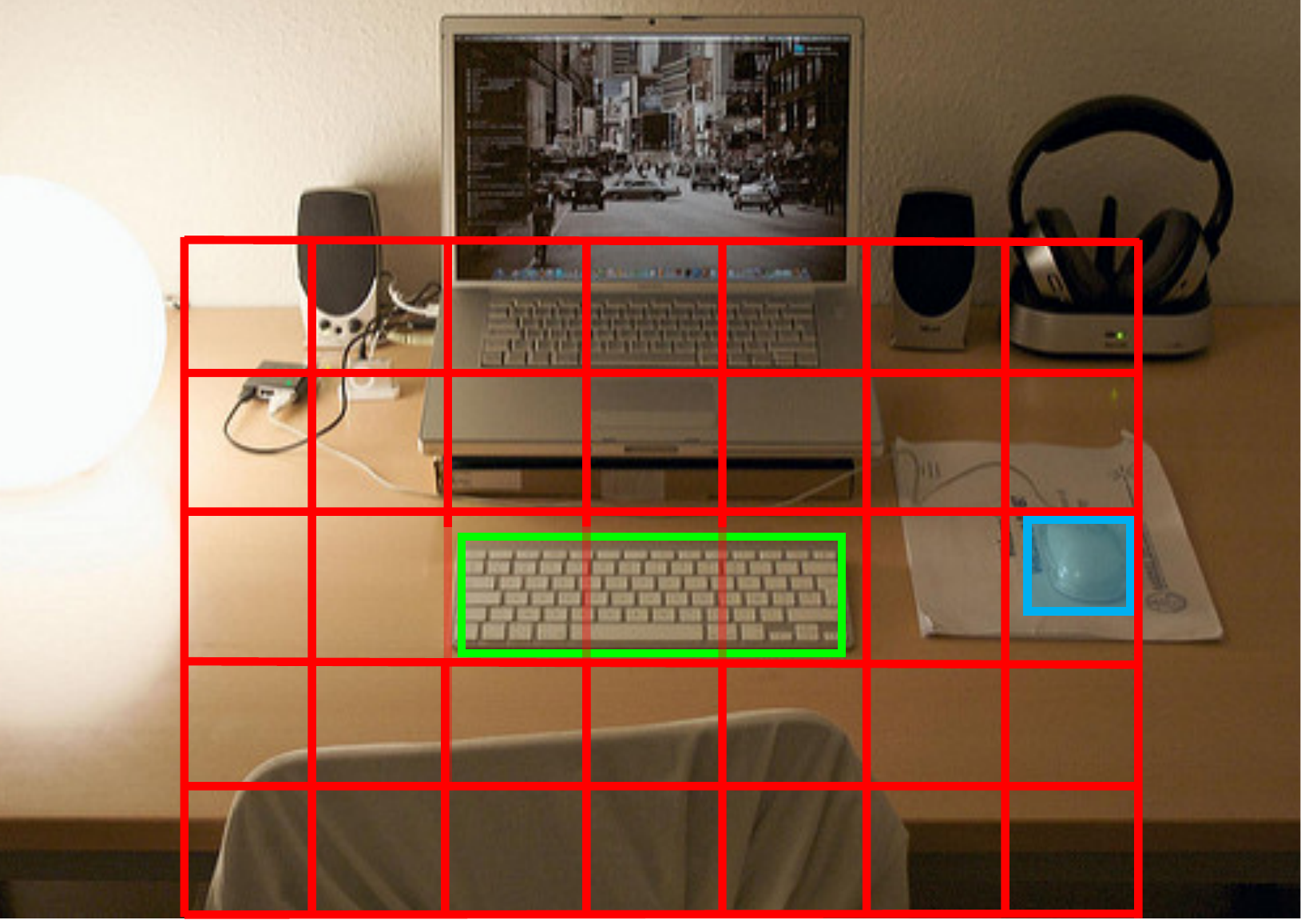}
	\end{center}
	\caption{The spatial context of a detection (in green) is determined relative to its location and height. In this case, the value of the contextual relation ``bin [0,3] has mouse'' is \emph{true}, where [0,0] represents the central bin.}
	\label{fig:spatial_context}
\end{figure}

Specifically, we experiment with co-occurrence relations (\eg $ctx_i$ indicates for detection $x_i$ whether a mouse exists in the image without overlapping $x_i$), spatial relations (\eg $ctx_i$ indicates whether a mouse exists in a specific image location relative to $x_i$), random context ($ctx_i$ is random binary number), or no context ($ctx_i=0$ for any $i$). Spatial relations are calculated in a manner invariant to object size and image location by examining the center point of each object in a reference frame centered on $x_i$ with bins whose size is determined as a factor of the height of $x_i$ (see Fig. \ref{fig:spatial_context}). Since different combinations of these relations may be more meaningful, for each object category we also experiment with and/or pairs of the 50 most improving relations. More specifically, for binary relations $rel_1$ and $rel_2$ we include binary relations $rel_1 \land rel_2$ and $rel_1 \lor rel_2$.

We test the described contextual relations using detection results provided by the Faster R-CNN detector \cite{ren_etal_2015_NIPS, huang_2017_CVPR} over the COCO \cite{lin_etal_2014_ECCV} validation set containing 80 object categories. Detections are considered correct/true if they overlap a ground-truth object with an intersection over union (IoU) of more than 0.5. Toasters and hair dryers are ignored due to their very few appearances. We assume that the context of each detection is known, and of course, in a real-world setting the obtained AP will be lower than the calculated bound since the context will not be perfectly extracted. As a discretization scheme for function $g$, the base confidence values are split into 10 bins and the context values, that are all binary, are naturally split into 2 bins. 

For each object category and contextual relation we find the AP upper bound and define the best relation as the one with the largest upper bound. The maximal obtainable improvement is then defined as the difference between the best relation's AP upper bound and the upper bound without context ($ctx_i=0$). We generate a detailed report containing the best contextual relation per object category as well as the best relation's amount of improvement. In the interest of space, we present notable examples and then show data in aggregate form. The entire report is provided in the supplementary material. 

Two examples from the categories with highest obtainable improvement are hot-dog and suitcase, with improvements of 4.7 and 5.5 AP units respectively, while two of those with the least possible improvement are zebra and cat, with improvements of 1.5 and 1.6 AP units respectively. The best relations for these categories were ``image has a person and another hot dog'' for hot dog, ``image has another suitcase'' for suitcase, ``bin [0, -1] (left) has a zebra or bin [0, 2] (right) has a zebra'' for zebra, and ``image has a bowl or another cat'' for cat. Note that a detection's center defines bin [0,0] as shown in Fig. \ref{fig:spatial_context}. Interestingly, for 70\% of the categories the best relation consist of only co-occurrence information (without bin locations), and when employing a stricter localization criteria of 0.75 IoU (instead of 0.5), this ratio goes down to 45\%. Therefore, significant improvements can be gained simply by recognizing and employing the existence of other objects in the image, but unfortunately, this kind of information becomes less relevant when the number of localization errors increases.  

\begin{figure}
	\begin{center}
		\includegraphics[width=1\linewidth]{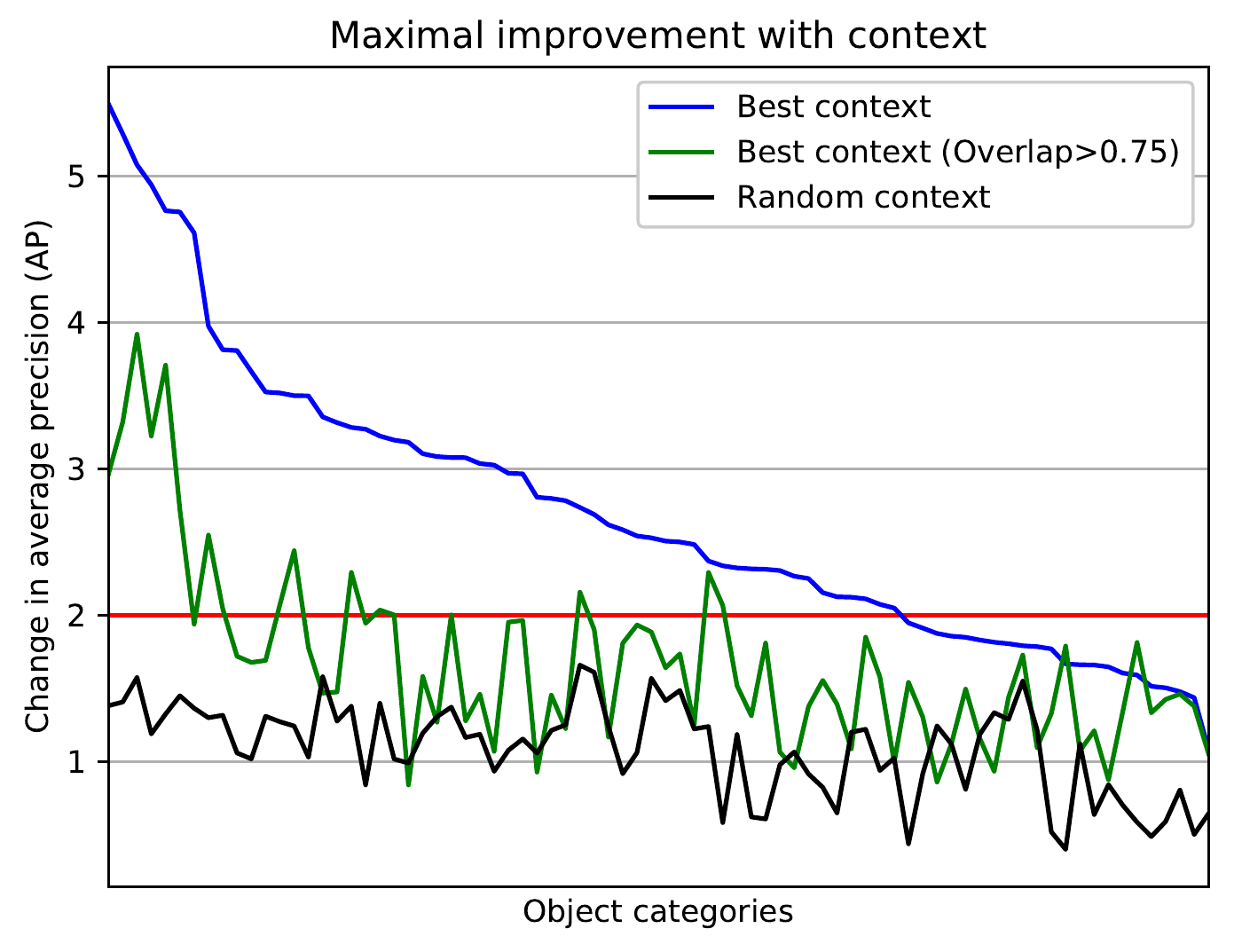}
	\end{center}
	\caption{The maximal AP improvement that can be obtained per object category with any type of context (blue and green), and with random binary context (black). X axis categories are sorted according to the blue curve, the blue and black curves are based on an overlap criteria of 0.5 IoU while the green curve represents a stricter case requiring 0.75 IoU.
	}
	\label{fig:max_imp}
\end{figure}

Aggregated results are presented in Fig. \ref{fig:max_imp}, providing more general insights into the use of context. The blue curve presents the maximal obtainable improvement per object category. As can be seen, the largest improvement for most categories is above the red line that marks an improvement of at least 2 AP units, but please remember that the calculated bounds are based on the best function $g$ and the context was assumed to be known. Therefore, for many object categories the maximal improvement that could be obtained in real conditions is expected to be lower or maybe even marginal at best. 

Due to the reliance on ground-truth information this method will always yield an improvement and may unjustifiably increase its measured amount. We therefore test the noise in this method by examining the improvement obtained by employing random context. More specifically, we show the average improvement of 10 trials with randomly generated binary context (black curve). As can be seen, the black curve is significantly lower than the blue curve, indicating that the present noise does not have a large effect.

We perform an additional experiment as in the blue curve in Fig. \ref{fig:max_imp} but this time with a stricter localization criteria of 0.75 IoU. The results can be seen in the green curve that is now significantly lower than the blue and quite similar to the black curve based on random context. The best obtainable improvement dropped dramatically in most cases, suggesting that object relations are not suitable when more accurate localization is required. In Sec. \ref{sec:separation} we show that context is simply unable of handling localization errors, explaining this drop.

In addition to COCO and Faster-RCNN, we repeat the same experiment with the SSD detector \cite{Liu_2016_ECCV} over the KITTI dataset \cite{Geiger_2012_CVPR} containing road scenes for autonomous driving. The largest improvements for cars, pedestrians, and cyclists are 0.7, 1.8, and 1.6 AP units, respectively. With localization criteria of 0.75 IoU the improvements reduce to 0.7, 1, and 1.2 AP units. Object relations are much less significant in this case, and still a similar drop is observed with stricter localization. A drop was not observed for cars, but that may be due to the already large number of strong localization errors (83\% of the errors at recall lower than 0.1 are due to localization). 

Finally, to test the ability of the heuristic suggested in Section \ref{sec:general_case} to maximize AP we attempt to reach a better AP by examining all bin orders. For each object category we revisit the relation that provided the highest AP upper bound and re-calculate the AP for any permutation of the bins. To allow such a costly procedure, we discretize the confidence into 5 bins instead of 10 and the context is still discretized into 2 since the tested context is binary. This requires to examine $10!$ bin combinations. Examining the largest AP upper bound found for each of the 78 object categories in the COCO dataset reveals that the AP based on our suggested heuristic is indeed maximal for most categories apart from several cases in which it provided a bound that is lower than the maximal by a negligible amount of at most 0.17 (where the AP is reported here between the range of 0 and 100). This result raises our confidence that the ranking of input bins according to the ratio of true to false detections well approximates the maximal AP.

\section{Analysis of Classification Capacity}
\label{sec:separation}

When given the results of a detector it is reasonable to wonder what exactly is required of context to improve them. Considering the role of context as strengthening true detections and weakening false detections may be generally correct, but this alone is uninformative. The method described above enables to analyze the gains that can be expected using different contextual relations, but it does little to explain \emph{why} context may be helpful in some cases and unhelpful in others. In this section we view the role of context as a means of distinguishing between true and false detections with similar base confidence. We then follow with an intuitive experiment to show just that by measuring the capacity of context to classify strong true and false detection.

The immediate effect of context on a set of detections becomes apparent when comparing Fig. \ref{fig:summary}a and Fig. \ref{fig:summary}c. In the former, detections of similar base confidence are grouped into bins, and the distribution of true and false detections inside the bins determine the AP. In the latter, an additional context dimension is introduced, further separating detections that were previously inseparable and allowing new confidence values to be assigned. All things considered, the role of context is thus to further separate detections with similar base confidence.

For a more intuitive understanding of why context helps, we analyze the capacity of context to classify a set of strong detections that consists of the same number of true and false detections.
Guided by the discouraging decrease in improvement for larger overlap thresholds (Fig. \ref{fig:max_imp}), we consider different types of false detections. We follow the analysis of detection errors by Hoiem \etal \cite{hoiem_etal_2012_ECCV} and define three error types -- localization errors, confusion with other classes, and confusion with the background. Localization errors are defined as detections with a label that matches the most overlapping ground-truth object with IoU overlap larger than 0.1. Confusions with other classes are detections for which the most overlapping ground-truth object overlaps by more than 0.1 but has a different label. Finally, confusions with background are false detections for which no ground-truth object overlaps by more than 0.1. 

To test the classification capacity of a contextual relation for some object category, we collect the $n$ most confident true detections and $n$ most confident false detections of one of the error types described above. The number $n$ is defined as the minimum between the available number of true and false detections (of the given error type). A binary relation separates the $2n$ detections into two groups containing a different number of true and false detections. Treating the context as a classifier, we label each detection as true if its group contains mostly true detections and false otherwise. To measure the wellness of classification we employ the accuracy of assigned labels and note that the minimal obtainable accuracy is 0.5 due to our use of ground-truth information in this analysis.

\begin{figure}
	\begin{center}
		\includegraphics[width=1\linewidth]{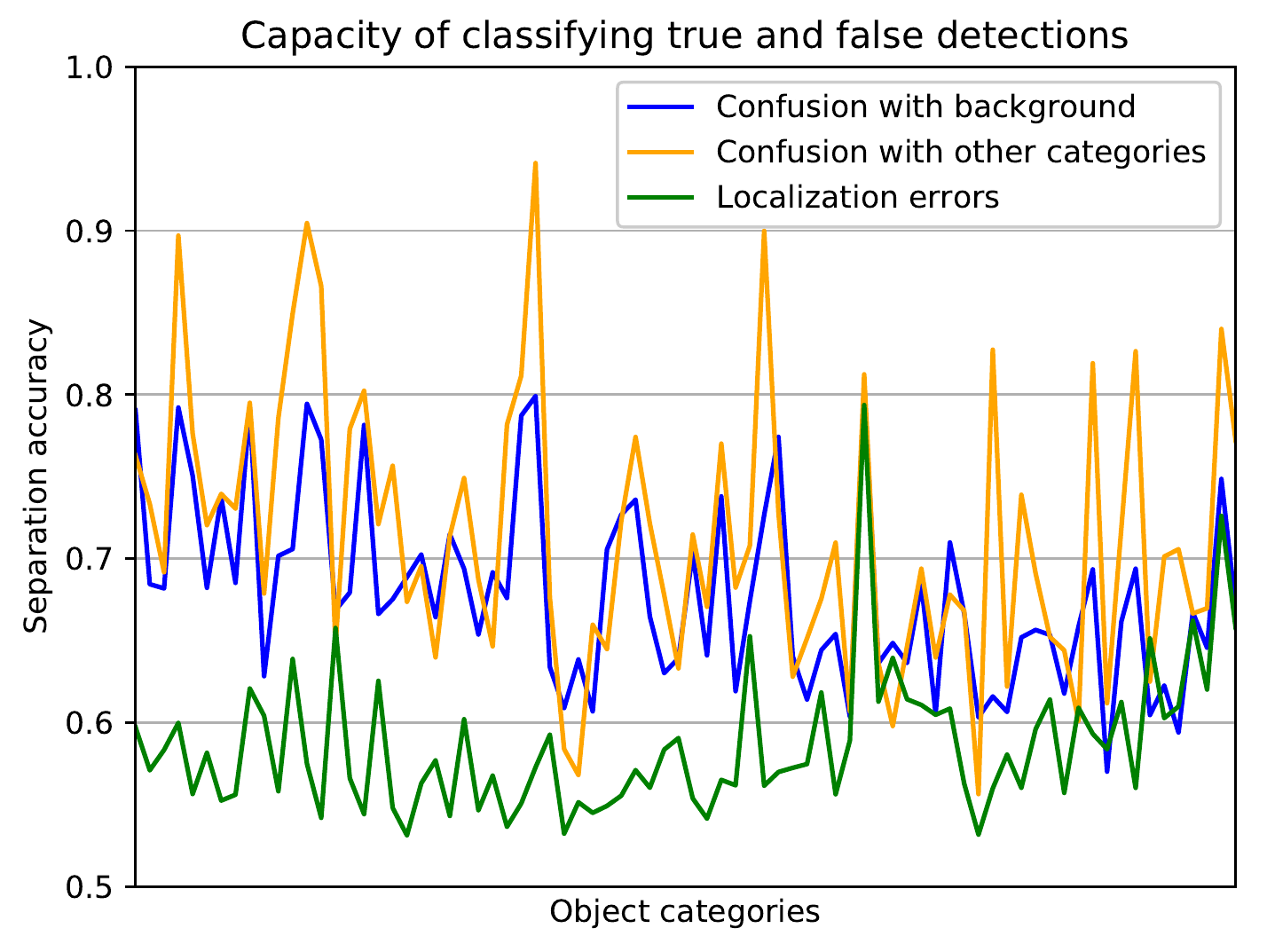}
	\end{center}
	\caption{The maximal classification capacity of context per object category. The context is used to classify detections as true or false, when only errors due to confusion with the background are employed as false detections (blue), when only confusions with other object categories are employed (orange), and when only localization errors are employed (green). X axis categories have the same order as in Fig. \ref{fig:max_imp}.
	}
	\label{fig:max_sep}
\end{figure}

We employ the contextual relations described in Sec. \ref{sec:bounding_exp} on Faster R-CNN over COCO. For each object category we report the maximal accuracy obtained by any of the relations (see Fig. \ref{fig:max_sep}). As can be seen, co-occurrence and spatial context is better at classifying, \ie distinguishing between true and false detections, when errors are only due to confusion with other categories than when errors are only due to confusion with the background. While it has some capacity to distinguish between true detections and localization errors, it is significantly lower and not much above the 0.5 threshold in most cases. This is likely to be the underlying cause for the significant drop in context-based improvement when increasing the localization threshold to an IoU of 0.75, as shown by the green curve in Fig. \ref{fig:max_imp}. It also affects the improvement at the standard IoU of 0.5 (as shown by the blue curve), or it might be the reason why context often hurts detection results (since strengthening true detections similarly affects localization errors). 

A similar trend is observed when repeating the experiment with the SSD detector over the KITTI dataset that contains cars, pedestrians, and cyclists. For these three classes respectively, the maximal classification accuracy is 65\%, 61\%, 64\% for confusions with the background, it is 75\%, 67\%, 58\% for confusions with other categories, and 57\%, 55\%, 54\% for localization errors.

\section{Discussion}
\label{sec:discussion}

Despite the large body of work done on the inclusion of context for object detection it still remains somewhat misunderstood. Our survey of results in Sec. \ref{sec:survey} clarifies the current state of research on the topic, showing that in many cases context does improve results and in many other cases the improvement is only marginal or even harmful. The theoretical analysis and empirical experiments in Sections \ref{sec:bounding_exp} and \ref{sec:separation} point to localization errors as one aspect that explains the low utility of context observed in many cases. This severely limits applications that require accurate localization. However, there may also be other applications for which accurate localization is not important, and reporting results that are based on an IoU of 0.5 or more may unexpectedly make the context appear non worthwhile. Of course, in some cases the ability of context to classify detections as true or false is rather low regardless or error type (Fig. \ref{fig:max_sep}), implying that for such object categories the tested context is simply not informative.

In addition to the reported AP, the inability to treat true detections and localization errors differently may have further discouraging properties. As context models are generally defined to increase the confidence of detections in favorable context, the same applies to localization errors. This problem, together with the issue that models are usually trained on different loss functions than AP, may be the reason for the large decrease in detection results that is often observed. For the same reason, it may be possible that the parameters learned by context models would provide a weaker context so as to avoid strengthening localization errors. In such cases, for applications that do not require accurate localization it may be preferable to train with a lower IoU (instead of just evaluating results with a lower IoU).

Another important point is the types of context employed. The experiments in this work focused on relations between objects via co-occurrence and spatial relations. However, there may be other kinds of additional information that could help when object relations cannot. For example, it is likely that the aspect ratio of a detection's bounding box or additional segmentation of its pixels may help correct localization errors \cite{divvala_etal_2009_CVPR}. These approaches too can be examined using the suggested analysis method but are outside the scope of this work.

\section{Conclusions}
\label{sec:conclusion}

Seeking to shed light on the use of context for object detection we have suggested a method for finding the function that combines contextual relations with standard detection results so as to maximize the detection score. Using this method we are able to show which relations are not informative, and to point to those that are more worthwhile to pursue and to the object categories that benefit the most. Further experiments highlight that a reason for the often discouraging results of employing context is its inability to handle localization errors, thus limiting the possibilities for improvement when confident localization errors are abundant. As a general guideline, the context can provide a significant improvement depending on the type of errors and when it is different for true and false detections with similar base confidence. Finally, we invite researchers to employ the tools developed here to analyze the improvement they can expect by incorporating context, and the contextual relations that provide it, in order to improve detection results.

\section*{Acknowledgments}
This research was supported in part by Israel Ministry of Science, Technology and Space (MOST Grant 54178). We also thank the Frankel Fund and Cyber Security Research Center, both at Ben-Gurion University of the Negev, for their generous support.


{\small
   \bibliographystyle{ieee}
   \bibliography{journal_names,all,ehud_cvpr2018}
}

\end{document}